\documentclass{article}
\usepackage[utf8]{inputenc}
\usepackage{geometry}

\usepackage{graphicx}
\usepackage[dvipsnames]{xcolor}
\usepackage{caption}
\usepackage[labelfont=bf]{subcaption}
\usepackage{cite}
\usepackage{amsmath}
\usepackage{amssymb}
\usepackage{url}
\usepackage{cmap}
\usepackage{hyperref}
\hypersetup{%
   colorlinks = true,
   linkcolor = uppercolor,
   citecolor = uppercolor,
   urlcolor = lowercolor}

\definecolor{uppercolor}{HTML}{0057B8}
\definecolor{lowercolor}{HTML}{FFD700}

\newcommand{\tr}{^{\top}}

\def\va{\mathbf{a}}
\def\vs{\mathbf{s}}

\def\vn{\mathbf{n}}

\def\vw{\mathbf{w}}
\def\vx{\mathbf{x}}
\def\vy{\mathbf{y}}

\def\mA{\mathbf{A}}

\def\mW{\mathbf{W}}
\def\mX{\mathbf{X}}

\def\mSigma{\mathbf{\Sigma}}

\newcommand{\RR}{\mathbb{R}}

\title{Towards physiology-informed data augmentation for EEG-based BCIs}
\author{Oleksandr Zlatov\\
    Neurotechnology Group\\
    Technische Universität Berlin\\
    \url{oleksandr.zlatov@tu-berlin.de}\\
    \and 
    Benjamin Blankertz\\
    Neurotechnology Group\\
    Technische Universität Berlin\\
    \url{benjamin.blankertz@tu-berlin.de}}
\date{\today}

\begin{document}

\maketitle

\begin{abstract}
Most EEG-based Brain-Computer Interfaces (BCIs) require a considerable amount of training data to calibrate the classification model, owing to the high variability in the EEG data, which manifests itself between participants, but also within participants from session to session (and, of course, from trial to trial). In general, the more complex the model, the more data for training is needed. We suggest a novel technique for augmenting the training data by generating new data from the data set at hand. Different from existing techniques, our method uses backward and forward projection using source localization and a head model to modify the current source dipoles of the model, thereby generating inter-participant variability in a physiologically meaningful way. In this manuscript, we explain the method and show first preliminary results for participant-independent motor-imagery classification. The accuracy was increased when using the proposed method of data augmentation by 13, 6 and 2 percentage points when using a deep neural network, a shallow neural network and LDA, respectively.
\end{abstract}

\section{Introduction}
Brain-computer interfacing~(BCI) is a technology that allows to extract time-resolved information about the mental state of a user from their brain signals. This information can be used to translate the user's brain activity into commands that can be used for communication or control of external devices \cite{clausen2017help}. There are also use cases of BCI technology apart from communication and control, see \cite{blankertz2016berlin,krol2022towards}.

State-of-the-art BCI approaches use data-driven methods to train classification or regression models from existing training data. This (labelled) training data needs either to be recorded from the respective participant (participant-specific classification), or a database recorded from many participants performing the very same task (participant-independent classification) is required. In both cases, it is desirable to reduce the amount of training data. In particular, classification models with high complexity like (deep) neural networks are data hungry, i.e., require a relative large amount of training data to work properly.

Different approaches were proposed to reduce the required amount of calibration data. They can be grouped into several categories~\cite{lotte_signal_2015, wu_online_2017}. \textit{Regularization} methods increase the robustness of models. For example, shrinkage~\cite{lotte_learning_2010} is widely used in BCI calibration, providing a regularized estimate of the covariance matrices. Some methods use apriori \textit{physiological information} to decide on features or channels that are likely to be useful and, thus, lead to the reduction of the calibration data requirement. In \textit{adaptive classifiers}, parameters are incrementally re-estimated and updated over time as new EEG data become available. Unsupervised adaptation methods~\cite{vidaurre2010toward,huebner2018unsupervised} are especially useful as they learn from unlabeled data gained from the actual usage of BCIs. \textit{Transfer learning} techniques~\cite{jayaram_transfer_2016, wu_active_2018, kindermans_transferring_nodate} 
use data from the same or similar tasks recorded from current or different subjects to improve the performance. In \textit{co-adaptive calibration} methods~\cite{vidaurre_co-adaptive_2011, faller_co-adaptive_2014} the mental strategy of the user and the algorithm of the BCI system are jointly optimized. Typically, the user is provided with feedback and the classifier model is continuously adapted. 

Another promising direction of research is \textit{artificial data generation}~(ADG), which also can be applied for data augmentation and calibration time reduction. Even simple methods based on ADG can be used to reduce calibration time~\cite{lotte_signal_2015}. With a focus on oscillatory activity-based BCIs, \cite{lotte_signal_2015} reviews calibration time reduction and suppression methods, and proposes simple, fast and efficient tools based on artificial EEG data generation. Three proposed approaches~(signal segmentation and recombination in the time and time domain; signal segmentation and recombination in the time-frequency domain; and artificial trial generation based on analogy) are tested on three datasets: motor imagery, workload and mental imagery, improving the performance. In~\cite{krell_rotational_2017}, rotational distortions were proposed to generate EEG signals. Besides segmentation-recombination and geometric transformation strategies, recent generative adversarial networks~(GANs)~\cite{goodfellow_generative_2014} and their improvements~\cite{arjovsky_wasserstein_nodate}, that are being successfully used in image generation, were adapted to produce EEG data~\cite{hartmann_eeg-gan:_2018, ozdenizci_transfer_2018, luo_eeg_2018, aznan_simulating_2019}. Even though these methods show promising results, generating realistic multi-channel EEG trials for data augmentation is currently not possible.

Here, we propose a method to augment existing data sets with re-generated data which fuses a model-based and a data-driven approach. It takes advantage of knowledge about neurophysics and geometry of the head. The idea is to do EEG dipole source localization with spatio-spectral decomposition~(SSD)~\cite{nikulin_novel_2011}, apply some perturbations to the dipole sources~(shifting and changing its orientation), and re-generate new data using a forward head model. This way, we add physiologically plausible variability in the data that can be exploited by classification models and allows them to generalize from little data in a reasonable way for across-participant transfer.

\section{Material and Methods}

\subsection{Material}
\label{material}

We evaluate the proposed method on a large-scale BCI study with motor imagery as control paradigm. Several papers analyzing the data set have been published, e.g., \cite{blankertz2010neurophysiological,hammer2012psychological,sannelli2019large}. The study was approved by the Ethical Review Boards of the Medical Faculty, University of Tübingen. 80 participants, who did not take part in any BCI study before, took part in the study, 39 male, 41 female (aged 29.9±11.5y, with a range of 17-65).

In the present analysis, we consider only the runs in which participants performed kinesthetic motor imagery \cite{neuper2005imagery} according to visual cues in offline mode, i.e., without BCI feedback and made the following two restrictions:

(1)~\emph{Restriction to motor imagery of the hands:}
While three types of motor imagery were used in the study, we consider only the two classes of \emph{left-hand} and \emph{right-hand} motor imagery. The reason for this restriction is the following: The third class of motor imagery involves either one \emph{foot} or of both feet, according to the participant's preference. Different from the hands, the effect of motor imagery of a foot is fundamentally different between participants. Few have a decrease in the amplitude of the sensorimotor rhythm (SMR) over the foot area (which would be expected and analogue to the effect of hand imagery), many have an increase of the SMR amplitude over the hand areas (probably due to inhibition, \cite{neuper2001event}), and some have an increase in rhythmic activity of the foot area, often in the beta range. These largely different correlates of foot imagery between participants makes learning participant-independent models more complex. This is a challenge that is not addressed in this first evaluation of the proposed method.

(2)~\emph{Restriction to participants with good baseline classification:}
Moreover, we included only data from those participants, for whom the (participant-specific) \emph{left-} vs.\@ \emph{right-hand} motor imagery classification yielded an accuracy of at least 80~\% (with the baseline CSP + LDA approach, \cite{blankertz2010neurophysiological}). While gaining better classification accuracies for participants in which baseline methods fail is an important goal of method development, it is not the primary goal of the method proposed here. We want to demonstrate that physiology-informed data set generation helps to obtain similar classification results with less training data. Applying this restriction, we retained 18 participants.

Brain activity was recorded from the scalp with multi-channel EEG amplifiers (BrainAmp DC by Brain Products, Munich, Germany) using 119 Ag/AgCl electrodes (reference at nasion; manufacturer EasyCap, Munich, Germany) in an extended 10–20 system, and sampled at 1000 Hz with a band-pass filter of 0.05 Hz to 200 Hz. For the present analysis, which is concerned with frequencies below 30~Hz, signals have been subsampled at 100~Hz. From each participant, 75 trials of each motor imagery class (\emph{left-hand}, \emph{right-hand}, and \emph{foot}) has been recorded in pseudo random. Each trial started with the presentation of a fixation cross for 2~s, which was then replaced for 4~s by an arrow pointing to the left, right or downwards to cue the respective motor-imagery task. Subsequently, the arrow disappeared, and a blank screen was shown for 2~s. After that, the next trial started. More details about the experimental setup can be found in \cite{blankertz2010neurophysiological}.

\subsection{Methods}
\label{sec:method}
Our method is based on the physiological information about brain. The signals that are acquired as EEG are mainly generated by pyramidal neurons in the cortex, which are oriented perpendicular to the cortical surface. The areas representing a certain function, like motor control of the left hand, are in similar locations across individuals. However, notable differences exist and, in particular due to the folding of the cortical surface, the orientation of the respective neurons is different. This can lead to substantial differences in the projection of the signals to the scalp. While the resulting large inter-participants differences observed in the EEG patterns pose a considerable challenge for participant-independent classifiers, we can also exploit this fact to modify data of one participant to mimic data of another (imaginary) participant. More specifically, we use a source decomposition method on the given data, localize the dipoles of the current sources, randomly change location of those dipoles, can recombine the source signals using a forward model with the modified dipoles to new natural-like EEG data. This idea can be used to augment a data set of, say 10 participants, with re-generated data of further 40 (imaginary) participants. This way of augmenting small training sets with natural variability should allow a more robust training of complex models.

\subsubsection{The linear model of EEG}
Our method relies on the general linear model of EEG \cite{parra2005recipes} which assumes that the measured EEG signals $\vx(t)\in\RR^c$ are a linear superposition of the source signals $\vs(t)\in\RR^d$
\begin{equation}
    \vx(t) = \mA \vs(t) + \vn(t)
    \label{eq:forward_model}
\end{equation}
where $\mA\in\RR^{c\times d}$ is the propagation matrix of the forward model with $c$ being the number of channels and $d$ being the number of modelled sources, and $\vn(t)$ represents the noise (brain components not captures by the model and external artifacts), often assumed to by Gaussian with zero mean. Each column of $\mA$ is a \emph{spatial pattern} which defines how the corresponding source signal is projected to the EEG channels. The counterpart is the backward model, which describes the decomposition of EEG signals with spatial filters:
\begin{equation}
    \vy(t) = \mW\tr \vx(t)
    \label{eq:backward_model}
\end{equation}
where each column of $\mW\in\RR^{c\times d}$ (resp.\@ each row of $\mW\tr$) is a spatial filter that extracts one component from the EEG. Given a matrix of spatial filters $\mW$ (backward model) for data $\vx(t)$ (e.g., obtained by a signal decomposition method), one can calculate the corresponding matrix of spatial patterns (forward model) by \cite{haufe2014interpretation}
\begin{equation}
    \mA = \mSigma_{\mX} \mW (\mW\tr \mSigma_{\mX} \mW)^{-1}
    \label{eq:filtertopattern}
\end{equation}
For invertible $\mW$, this implies $\mA = (\mW\tr)^{-1}$.

\subsubsection{Data generation}
\label{sec:data_generation}
\begin{figure}[!ht]
\centering
\includegraphics[width=\linewidth]{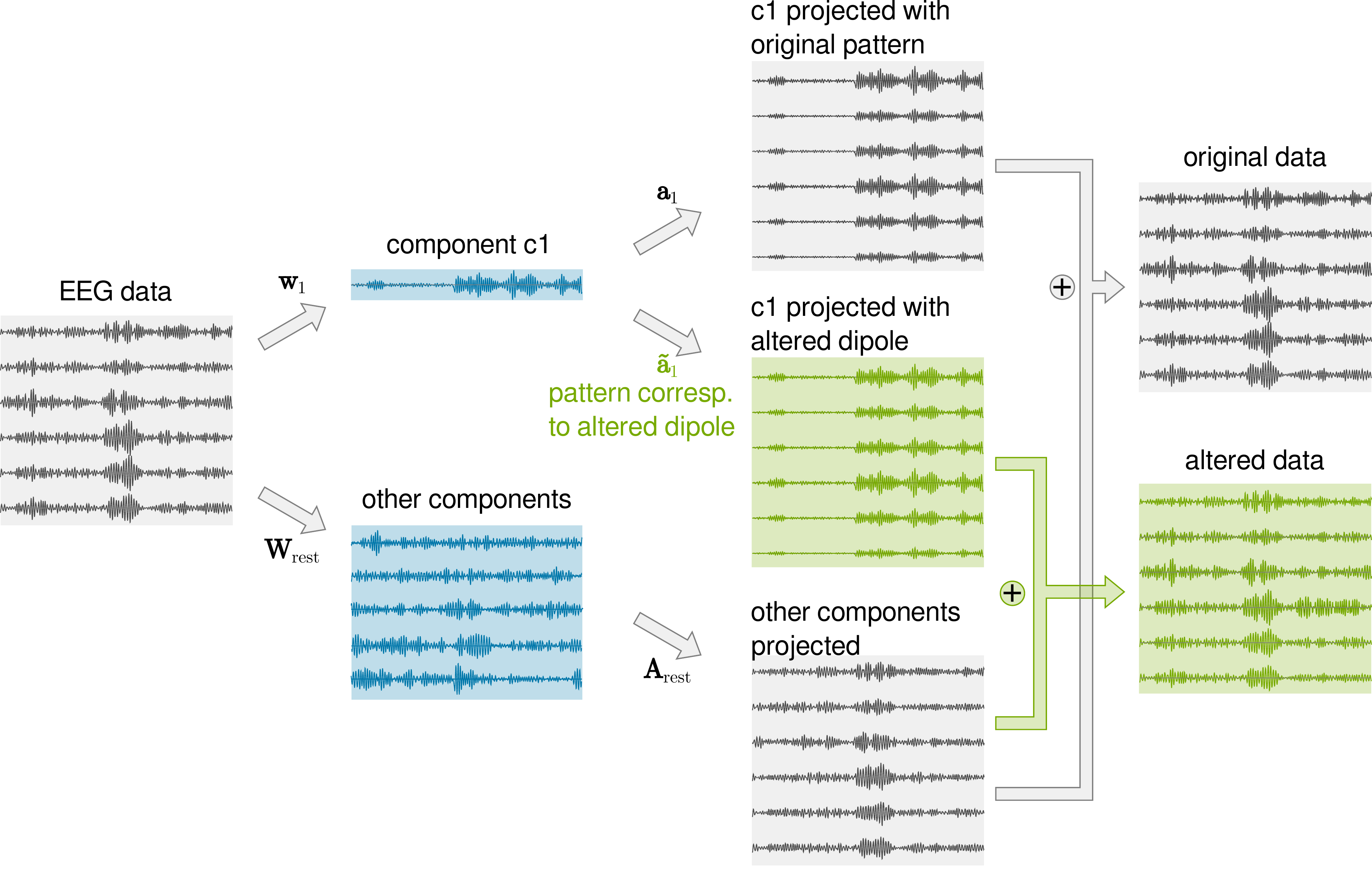}
\caption{Generating data with augmented patterns. The measured signals are decomposed with a backward matrix $\mW$. One component (here called c1) is selected. Its source activity is extracted with the spatial filter $\vw_1$. As an alternative to the corresponding pattern $\va_1$, an augmented pattern $\tilde{\va}_1$ is determined by changing the fitted dipole. This process is detailed in Fig.~\ref{fig:dipole_modification}. Mixing the projected signals of the augmented pattern with the projected signals of the remaining components yields the generated data. The regeneration of the original signals (in the upper right) is just included in the figure for illustration of the method. Note, that also dipoles of several components can be altered, see text.}
\label{fig:generation}
\end{figure}

The data generation procedure consists of the following steps:
\begin{enumerate}
    \item Signal decomposition: We use the Spatio-Spectral Decomposition (SSD) method \cite{nikulin_novel_2011} to decompose the EEG signal into components. SSD determines a matrix of spatial filters $\mW$ by maximizing the ratio of power in the frequency band of interest and the power in flanking frequency bands in the extracted components. For the application to motor imagery data, we chose 8-13~Hz as the band of interest and 5-8~Hz and 13-16 Hz as flanking frequencies. The corresponding patterns $\mA$ are determined according to eqn.~\eqref{eq:filtertopattern}.
    \item Source localization. The MUSIC algorithm \cite{mosher1992multiple} is used to find a dipole fit for a given SSD pattern using a head model. The output is the location (voxel of the head model), orientation (normalized dipole moment) of the best fitting dipole, and corresponding source signal.
    \item Dipole modification. The dipole position is shifted to another voxel of the head model in its neighborhood, and the corresponding (`augmented') pattern is determined. To generate data for other imaginary participants, the dipole is shifted to different voxels.
    \item Data generation. The augmented pattern is used to project the source component using eqn.~\eqref{eq:forward_model}, see also Fig.~\ref{fig:generation}.
    It is possible to augment several SSD patterns. In this case, we apply steps 2 and 3 separately for each SSD pattern and sum up the resulting signals.
\end{enumerate}

Since the potential positions for dipole shifts are restricted by the grid of the head model, we cannot shift the dipole exactly to a predefined distance. To shift the dipole $N$ times, we select $N$ closest voxel to the original location, that have at least a predefined distance.

There are several options for selecting the components that are modified. One could use only the strongest SSD components, or only the strong components with dipoles in the sensorimotor area. One could also take those components which can be fitted well with a single dipole. In this analysis, we opted for altering all components (that a provided by SSD).

\begin{figure}[!ht]
\centering
\includegraphics[width=0.8\linewidth]{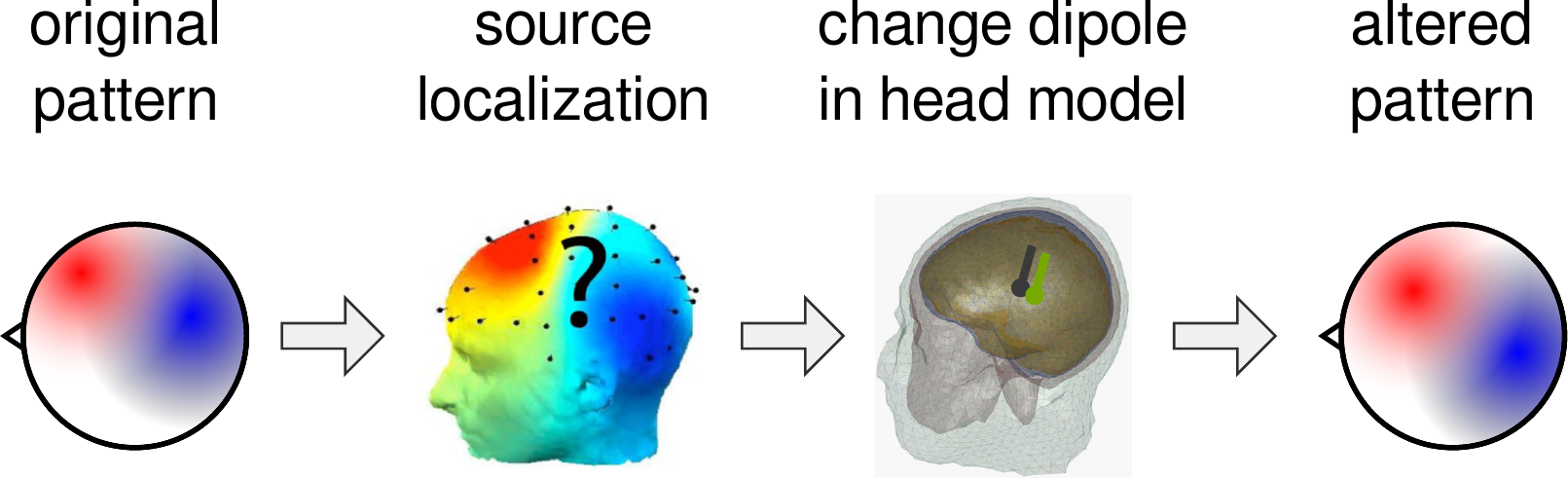}
\caption{Process of dipole manipulation. EEG signals are decomposed by SSD (backward model, eqn~\eqref{eq:backward_model}). For each component, MUSIC is used to find a single dipole fit. The location of the dipole is modified in a three-compartment head model. This yields a pattern that corresponds to the changed dipole, which can be used to generate new signals, see Fig.~\ref{fig:generation}.}
\label{fig:dipole_modification}
\end{figure}

\subsection{Classification}

\subsubsection{Preprocessing}
\label{sec:classification:preprocessing}
To get more focal signals, we applied 19 Laplacian (spatial) filters and a band-pass filter pf 8 to 13~Hz to the EEG signals. We extracted epochs in a time interval of 3.5~s duration starting 1000~ms after the visual cue from each trial.

\subsubsection{Linear Discriminant Analysis (LDA)}
For the classification with LDA, we calculated the variance within each trial and applied the logarithm. This resulted in 19-dimensional features. We used LDA with shrinkage of the covariance matrix \cite{vidaurre2009time,blankertz2011single} with the Ledoit-Wolf estimator \cite{ledoit2004well} of the shrinkage parameter using the method presented in \cite{schafer2005shrinkage}.

\subsubsection{Network architectures}

We used two convolutional neural networks, introduced in \cite{schirrmeister_deep_2017}, to evaluate the proposed method. Both we applied to the preprocessed epochs as explained above, see Sec.~\ref{sec:classification:preprocessing}.

\textbf{Deep ConvNet.}
The deep ConvNet contains four blocks of convolution and max pooling. The first block is split into temporal and spatial convolutions, which was shown to better handle the 2D-array input of EEG data, and lead to a better performance \cite{schirrmeister_deep_2017}. This block is followed by three standard convolution-max-pooling blocks and a dense softmax classification layer. The exponential linear units (ELUs) are used as activation functions.

\textbf{Shallow ConvNet.}
The first two layers of the shallow ConvNet are the temporal and spatial convolutions, as in the deep ConvNet. After that, a squaring nonliniarity, a mean pooling and a logarithmic activation functions are applied. These steps imitate the log-variance computation, which is used, for example, in CSP based classification \cite{blankertz2007optimizing}. Finally, the dense softmax classification layer is applied. 

The details of the architectures can be found in the original paper \cite{schirrmeister_deep_2017}. We had to decrease the length of the fourth convolutional filter (from 10 to 3) considering the length of trials in our data set, and adjust the final softmax layer for binary classification in our case. We followed the recommendations about the values for the hyperparameters, including learning rates, batch sizes and choice of optimizers. The preliminary grid search has shown that they are also close to optimal for our dataset.

\subsection{Evaluation}
We evaluated the method on the motor imagery dataset described in section~\ref{material} for the goal of participant-independent classification. That means, for probing the classification performance for a particular participant, we trained a classifier only on data of other participants. To create a setting of a very limited training set, we used only a small number of trials $k$ from each participant for training. In the presented analysis, $k=15$ was used.

Then we considered the number of modifications $N$. $N=1$ means that only the original data was used. For $N>1$, each component was modified in $N-1$ different ways, as described in section~\ref{sec:data_generation}. Accordingly, for each available participant, data of $N-1$ imaginary participants have been generated. And the augmented data set has $N$ times the size of the original data set. A small exploration suggested that $N=5$ works well for LDA, while $N=10$ is a good choice of the neural networks.

After this, we performed leave-one-subject-out cross-validation to check if adding generated data to the training set improves the performance of classification. For this, we compared two settings. First, we trained the classifier on $k$ original trials of $17$ participants, and tested on $150$ original trials of the target participant. Second, for all $17$ training participant, we added $k(N-1)$ generated trials to the training dataset, while still testing on $150$ original trials of the target participant. Better accuracy in the second setting indicates that adding generated data improves classification performance.

\section{Results}
\begin{figure}[!ht]
  \begin{subfigure}[b]{0.32\textwidth}
      \includegraphics[width=\textwidth]{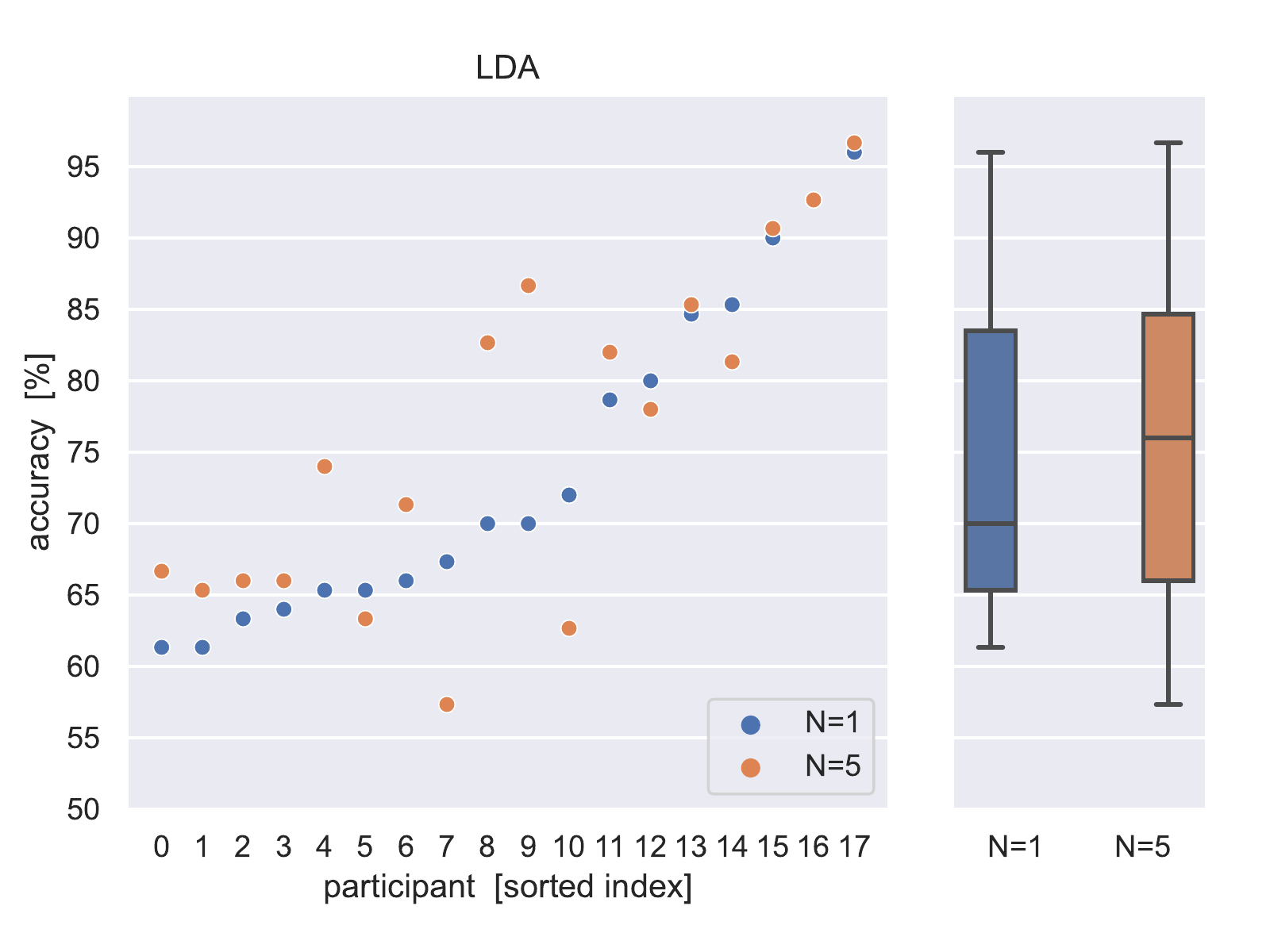}
      \caption{LDA}
  \end{subfigure}
  \hfill
  \begin{subfigure}[b]{0.32\textwidth}
      \includegraphics[width=\textwidth]{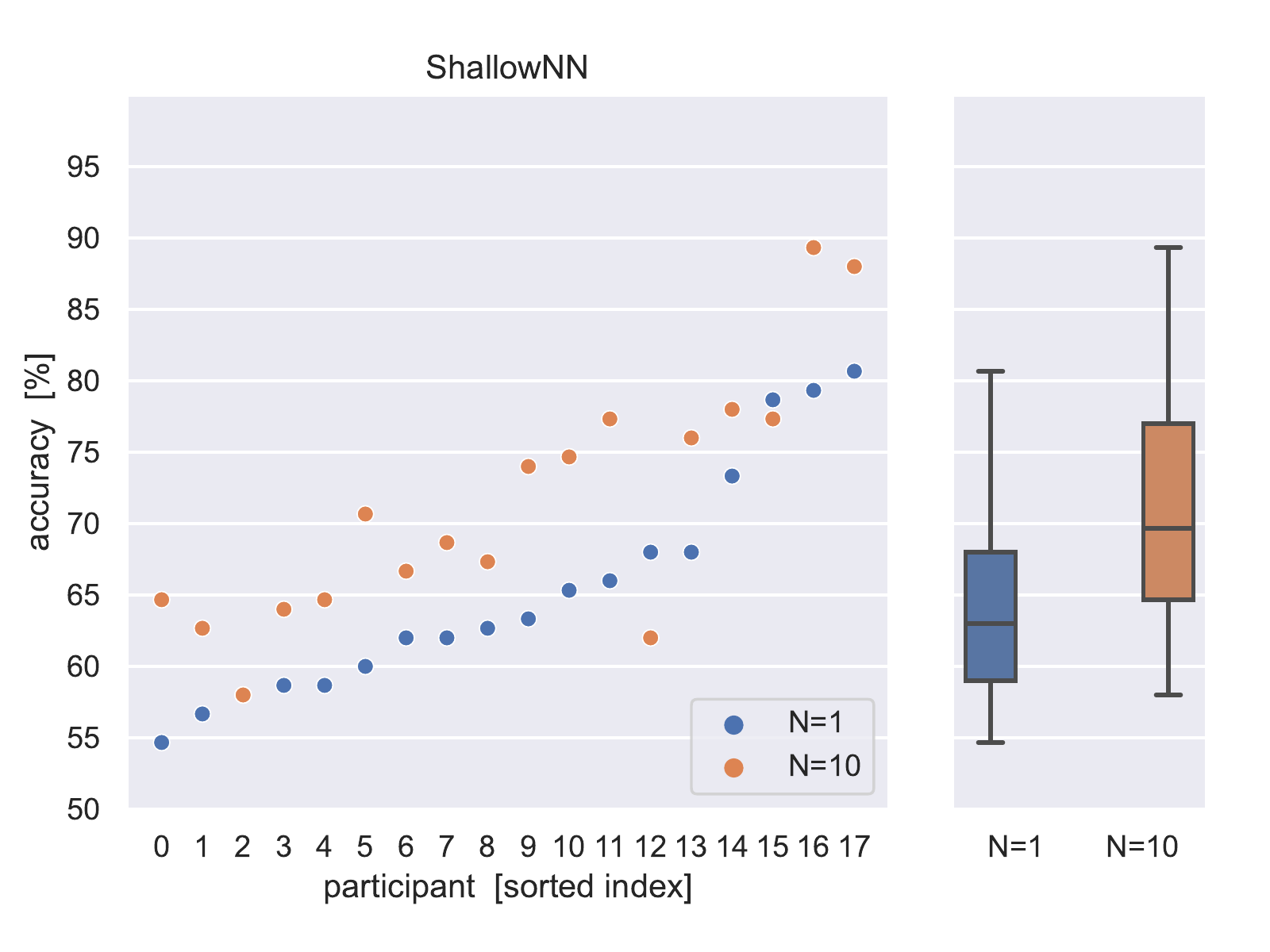}
      \caption{Shallow Network}
  \end{subfigure}
  \hfill
  \begin{subfigure}[b]{0.32\textwidth}
      \includegraphics[width=\textwidth]{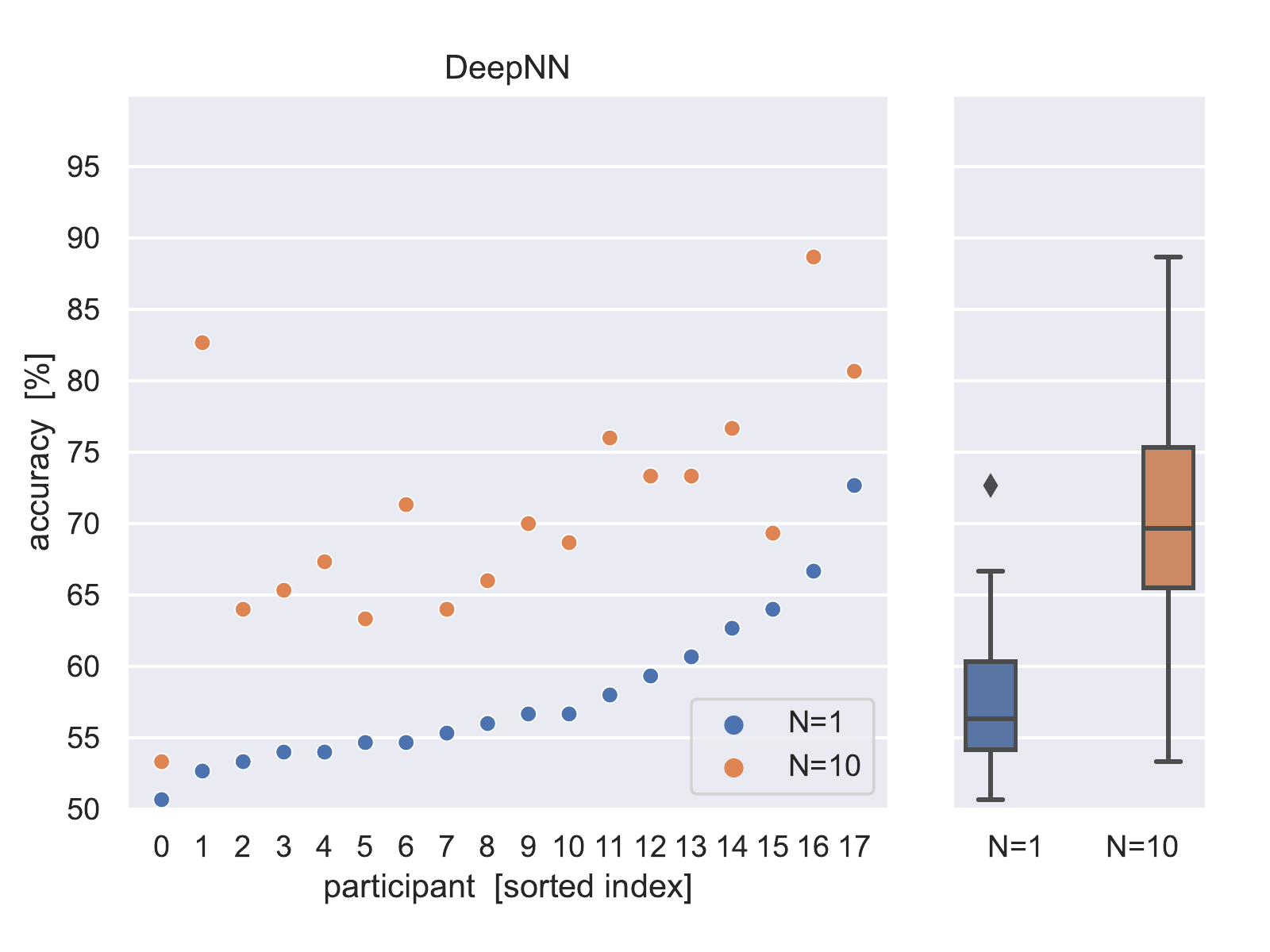}
      \caption{Deep Network}
  \end{subfigure}
\caption{Classification accuracy (percent correct) with different classification model. Results for $N=1$ refer to classification without data augmentation. Classification is done with participant-independent classifiers using a leave-one-participant-out validation scheme. Participants on the x-axis are sorted separately for each classifier regarding to $N=1$ results.}
\label{fig:results|participantwise-sorted}
\end{figure}

The results of participant-independent classification are shown in Fig.~\ref{fig:results|participantwise-sorted}. The results for $N=1$ correspond to baseline classification without data augmentation, while $N=5$ (LDA) and $N=10$ (neural networks) corresponds to data augmentation with adding data of $4$ resp.\@ 9 imaginary participants. The participants on the x-axis in each subfigure are sorted according to baseline classification individually for each classifier type. Baseline classification results are on average decreasing with increasing model complexity (from LDA to shallow to deep network). The gain in accuracy by data augmentation increases with model complexity.

\section{Discussion}
The results demonstrate the efficiency of the physiology-informed data generation. The accuracy increases for all investigated classifiers when augmenting the (small) training data with re-generated data. With the deep neural network, this is true for each individual participant, with the shallow network for almost all participants, and with LDA for the clear majority. The accuracy (percent correct) increased on average ($\pm$ standard error of the mean (SEM)) from $74.1\pm2.6$ to $76.0\pm 2.7$ for LDA, from $65.3\pm1.8$ to $71.3\pm2.0$ for the shallow network, and from $57.9\pm1.3$ to $70.8\pm1.9$ for the deep network. As expected, the accuracy in this small training data scenario goes down with the complexity of the classification model. And likewise, as hypothesized, the positive effect of data augmentation increases with model complexity. 

Still, albeit the boost in accuracy for the deep network of about 13 percentage points through data augmentation, the performance is still slightly inferior to the performance of LDA without augmentation. Complex networks have many degrees of freedom, and with small training sets it happens easily that they overfit to random correlation (noise) in the data. The goal of our data augmentation is to provide enough plausible variations of the signals such that complex models can be trained without overfitting. While the huge improvement of the neural network models show that our approach works in principle, it still does not work well enough to allow robust learning of complex models on our data sets.

There are a number of possible reasons. Either the networks are not adjusted well enough to the data at hand. Or the data augmentation introduces some artifacts in the data, e.g., in components that cannot be approximated well with single dipoles, which could reduce the utility of the additional data. Also, the current analysis tries to compensate also a missing inter-trial variability in the training data by adding plausible inter-participant variability, which might be possible only to a certain degree.

Future work will explore restrictions on which components are modified, consider further variations of the signals like shifting frequencies or changing the amplitude between components, and analyze how well the method can compensate when data from a smaller number of participants are available.

\subsection*{Funding}
OZ acknowledges support from the Research Training Group (RTG 2433) DAEDALUS (Differential Equation- and Data-driven Models in Life Sciences and Fluid Dynamics) funded by Deutsche Forschungsgemeinschaft (DFG).

\bibliographystyle{apalike}
\bibliography{bibliography}

\end{document}